\documentclass[10pt,twocolumn,letterpaper]{article}

\usepackage{iccv}
\usepackage{times}
\usepackage{epsfig}
\usepackage{graphicx}
\usepackage{amsmath}
\usepackage{amssymb}
\usepackage{graphicx}
\usepackage{multirow}
\usepackage[table,xcdraw]{xcolor}
\usepackage{dblfloatfix}
\usepackage{algorithm}
\usepackage{algpseudocode}
\usepackage{adjustbox}
\usepackage{booktabs}

\usepackage[pagebackref=true,breaklinks=true,letterpaper=true,colorlinks,bookmarks=false]{hyperref}

\iccvfinalcopy 

\ificcvfinal\fi

\begin{document}

\title{Prior based Sampling for Adaptive LiDAR}

\author{Amit Shomer\\
Tel Aviv University\\
{\tt\small amitshomer@gmail.com}
\and
Shai Avidan\\
Tel Aviv University\\
{\tt\small avidan@eng.tau.ac.il}
}

\maketitle
\ificcvfinal\thispagestyle{empty}\fi

\begin{abstract}

We propose SampleDepth, a Convolutional Neural Network (CNN), that is suited for an adaptive LiDAR. Typically, LiDAR sampling strategy is pre-defined, constant and independent of the observed scene.  Instead of letting a LiDAR sample the scene in this agnostic fashion, SampleDepth determines, adaptively, where it is best to sample the current frame. To do that, SampleDepth uses depth samples from previous time steps to predict a sampling mask for the current frame. Crucially, SampleDepth is trained to optimize the performance of a depth completion downstream task.
SampleDepth is evaluated on two different depth completion networks and two LiDAR datasets, \textit{KITTI Depth Completion} and the newly introduced synthetic dataset, \textit{SHIFT}. We show that SampleDepth is effective and suitable for different depth completion downstream tasks. Our code is publicly available\footnote{\url{https://github.com/amitshomer/SampleDepth}}.

\end{abstract}

\section{Introduction}

LiDAR is shaping to be an important ingredient in autonomous vehicles, as well as other AR/VR applications. As such, it is of interests to make it as efficient as possible.

There are several types of LiDAR and here we focus on scanning LiDARs, where the LiDAR scans the scene in order to measure the distance to different points in the 3D environment. The sampling pattern of LiDARs today is based on the hardware constraints of the device itself and are agnostic to the 3D scene. In this setting, the LiDAR produces a sparse set of depth measurements and extensive literature evolved over the years to develop depth completion algorithms~\cite{van2019sparse,li2020multi,ma2018sparse,park2020non}. These algorithms take the sparse depth map and the corresponding RGB image captured by a camera, to produce a dense depth map.

Clearly, increasing the number of depth samples will improve the reconstructed depth map. On the downside, it will prolong acquisition time and increase hardware cost. We wish to find a better sampling pattern that will achieve better depth completion results with the same sampling budget.

\begin{figure}[t!]
\begin{center}
\includegraphics[trim=0cm 7cm 0cm 0cm,width=\columnwidth]{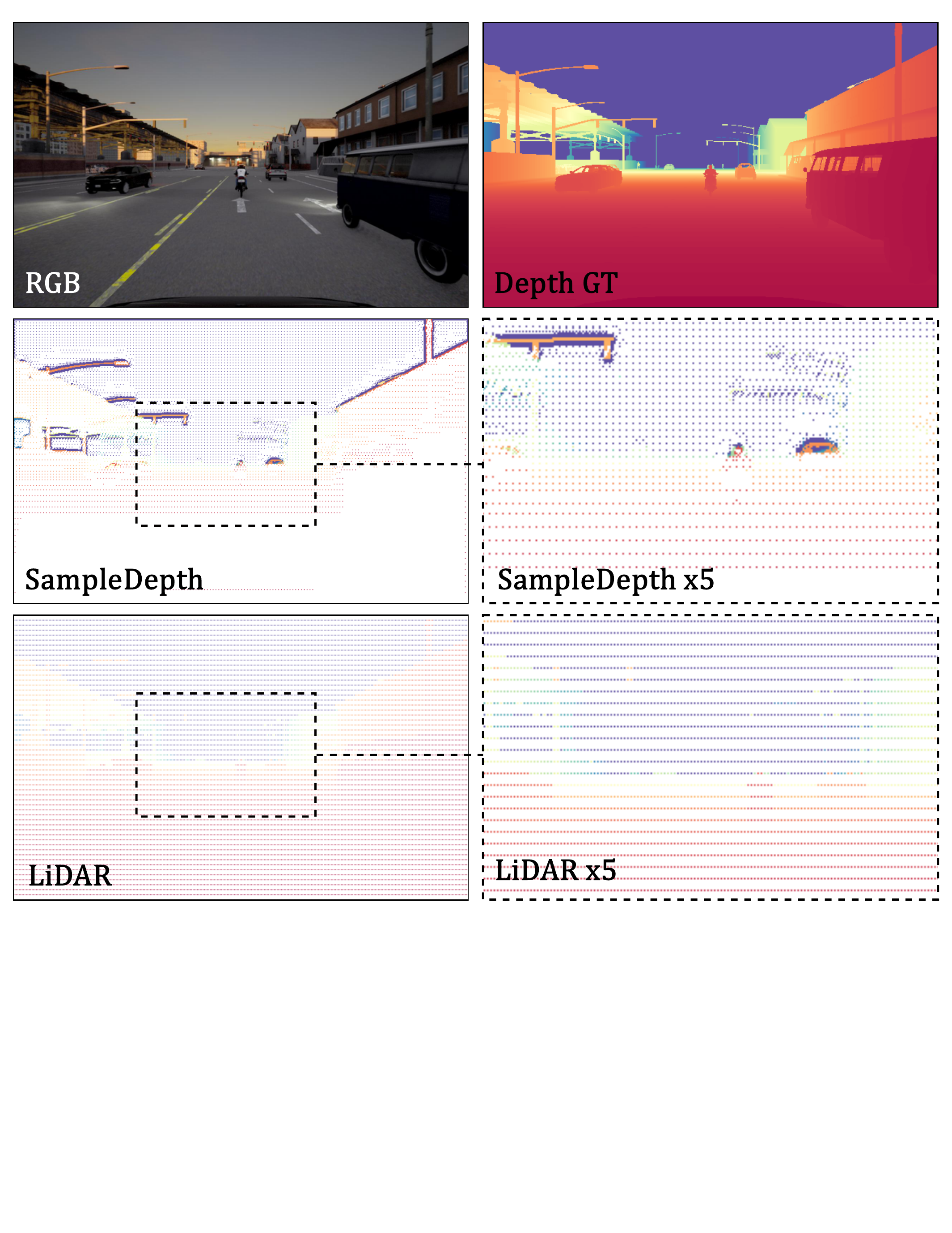}
\caption{{\textbf{Prior based sampling}. Our solution uses information from previous time steps as priors in order to predict the the position of the adaptive LiDAR sample mask for the current frame.} }
\label{fig:teaser}
\end{center}
\end{figure}

Towards this end, we consider three scenarios. The first, that serves as a baseline, is the {\em agnostic} sampling pattern that is in use today. In this setting, the sampling does not adapt to the scene. The second is a {\em fixed} sampling pattern that is based on learning a fixed sampling mask for the entire training set. For example, learning an optimal mask on a large collection of videos collected by a driving vehicle. The last is an {\em adaptive} sampling mask that is based on previous samples. This is the focus of this work. Figure~\ref{fig:teaser} shows the difference between agnostic sampling and adaptive sampling.

Our workhorse is a novel network, termed {\em SampleDepth}, that uses depth maps from previous time steps to determine the current sampling mask. Crucially, SampleDepth is trained subject to a target depth completion network, so it can be used in conjunction with any such network. Moreover, SampleDepth can be trained with one depth completion network and then used, as is, with another. The use of temporal data is a common practice in computer vision for a variety of tasks. However, to the best of our knowledge, we are the first to use it for adaptive LiDAR sampling.

We investigate two variants of the adaptive approach. In the first, we actually predict the current depth map from past measurements. In the second, we simply provide the past measurements and let the network implicitly learn from it.

We evaluate SampleDepth on two datasets, a real one and a simulated one, and find that it outperforms the existing agnostic sampling approach across a range of sampling rates. Furthermore, we show that SampleDepth can be used with different depth completion networks and, in fact, can be trained with one depth completion network and then used, as is, with another.

\section{Related Work}
\textbf{Depth completion.} The purpose of depth completion tasks is to take a sparse depth map and transform it into a dense depth map. Uhrig \etal~\cite{uhrig2017sparsity} suggests using convolutional neural networks (CNN) to complete sparse laser scan data. Since then, many studies~\cite{chodosh2019deep,ku2018defense,bai2020depthnet,eldesokey2018propagating} demonstrated that neural networks are capable of creating dense depth maps by taking only sparse LiDAR scans as an inputs. Other studies~\cite{ma2018sparse,park2020non,cheng2019learning,qiu2019deeplidar,van2019sparse} used also an RGB images to guide the depth completion process. 

Several studies have examined the effects of different depth map sparsity on the depth completion performance. Qiu~\etal~\cite{qiu2019deeplidar} uniformly sub-sample the raw LiDAR maps and checked the performance of their model against it. Karaman and Ma~\cite{ma2018sparse} randomly sampled the ground truth in order to increase robustness and increase training data size. Lu~\etal~\cite{lu2021sgtbn} use only a single-line of the LiDAR and an RGB image to generate dense depth map. 
\\\\
\textbf{Depth Sampling.} Recent works investigated a variety of sampling techniques for 3D points. The sampling of those points is intended to optimize various tasks, including classification, segmentation or reconstruction for a given number of sampling points. 
Dovrat~\etal~\cite{dovrat2019learning} proposed a learned task-oriented simplification which process raw point clouds. Lang~\etal~\cite{lang2020samplenet} extended this work by approximating the sampling operation with a differentiable nearest neighbor operator.

A number of studies focused on sampling 3D points that are projected into 2D maps.  These studies examined the possibility of obtaining better sampling patterns than what is achievable today with LiDAR hardware for the reconstruction task. To maximize the performance of disparity map reconstruction, Liu~\etal~\cite{liu2015depth} proposed a two-stage random sampling scheme. Wolf~\etal~\cite{wolff2020super} introduced an algorithm that uses super-pixels to divide the image plane, and then sample each super-pixel. In another work, by Bergman~\etal~\cite{bergman2020deep}, a deep neural network was used to sample at  very low sampling rates. Initially, they use grid sampling and then move those points to better locations using a vector flow fields.

Gofer {\em et al.}~\cite{gofer2021adaptive} use an ensemble of black-box predictors to determine the next point to sample. They do so by dynamically selecting the next point to sample within the current frame. Tcenov and Gilboa~\cite{tcenov2022guide} extend that work. First, they construct an uncertainty map by using a neural network guided by RGB images. Then, they create the final sampled map using Gaussian sampling. In the sampled process, both Gofer~\etal~\cite{gofer2021adaptive} and Tcenov~\cite{tcenov2022guide} utilize an iterative approach.

Taguchi~\etal~\cite{taguchi2023uncertainty} use iterative approach as well. Contrary to Gofer {\em et al.}~\cite{gofer2021adaptive}, who create a new uncertainty map every iteration using the depth completion task, they create an uncertainty map only once. Afterwards, a neural network module is used to update it in every iteration.

Without exception, all the sampling methods mentioned above are not using prior past frames, as we do.
\\\\
\textbf{Video prediction} is the task of inferring future video frames based on past frames. To address this issue, a number of approaches have been proposed. Some approaches~\cite{wang2022predrnn,hsieh2018learning, srivastava2015unsupervised,villegas2017learning,villegas2017decomposing} use recurrent neural networks (RNN) to make next frame prediction. 

Transform inputs are another popular method for synthesizing frames. Liang~\etal~\cite{liang2017dual} proposed a generative adversarial network (GAN) with joint optical-flow. Jiang~\etal~\cite{jiang2018super} learned a model to predict offset vectors in order to interpolate frames.
Liu~\etal~\cite{liu2019video} expanded this method by creating a spatially-displaced convolution module. The module learned the new value and the displaced location for each pixel, and used that to synthesize a new image.

Simpler approaches use a CNN to predict the next frame. A modified U-Net~\cite{ronneberger2015u} was used by Liu~\etal~\cite{liu2019video} to predict the next frame and to detect anomalies. Gao~\etal~\cite{gao2022simvp} achieved state-of-the-art results using CNNs trained with MSE loss.

\section{Method}
\begin{figure*}[h!]
    \centering
    \includegraphics[trim=1.3cm 18.2cm 0cm 0cm, width=\linewidth]{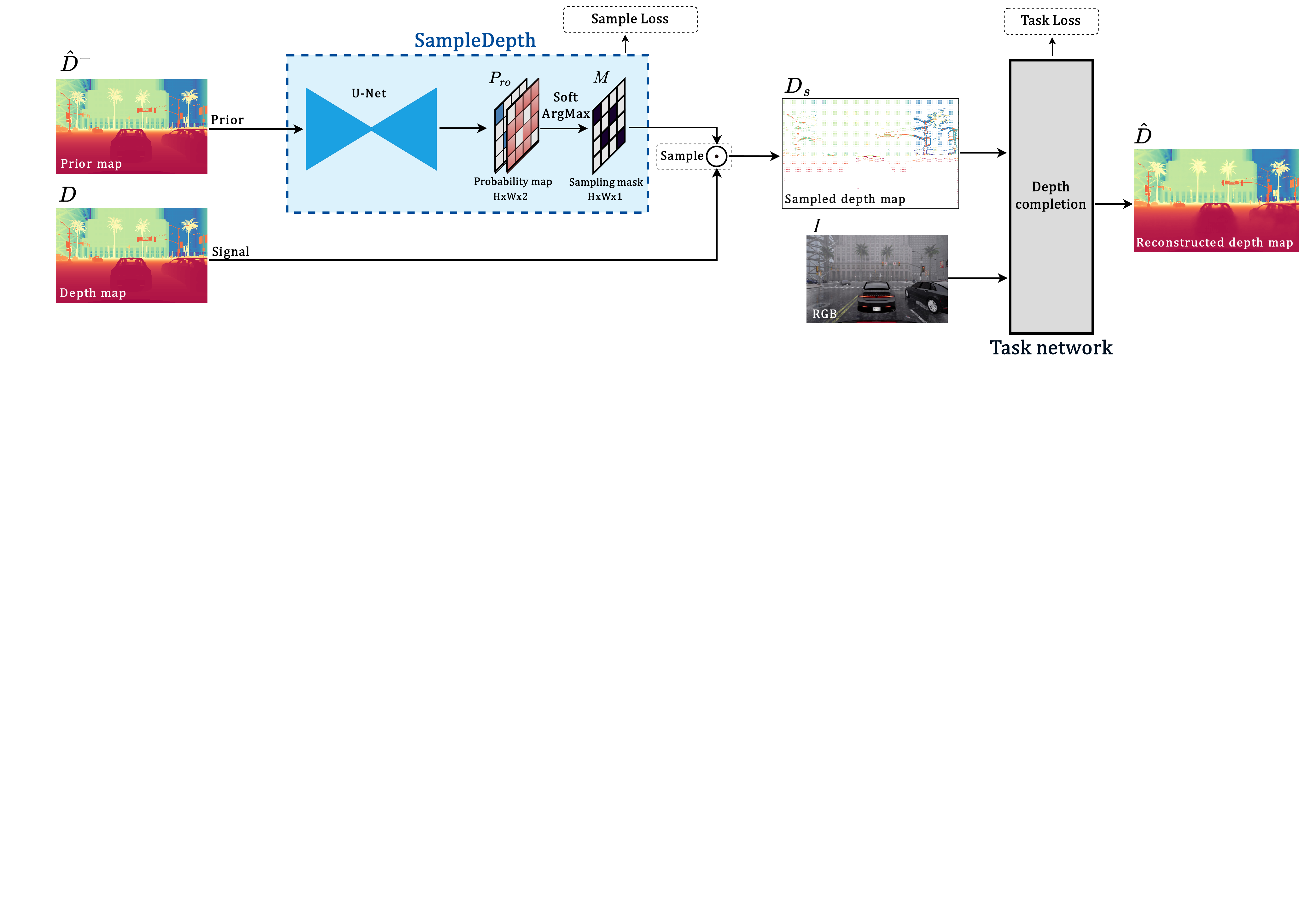}
    \caption{\textbf{The proposed SampleDepth sampling process.} The SampleDepth network takes the prior input and reproduces the sample mask as an output. In this example, the sampled signal and the prior are the same. In  section~\ref{results} we explore additional configurations of priors and signals.}
    \label{fig:training-regime}
\end{figure*}

A depth completion network $f$ takes an RGB image $I$ and a sparse depth map $D_S$ to create a dense depth map $\hat{D}$ that is as similar to the ground truth depth map $D$ as possible. The sampling pattern is governed by a sampling matrix $M$ with $k = |M|$ samples. That is, $D_S = D \odot M$, where $\odot$ denotes element-wise multiplication (i.e., the elements of $M$ are either $0$ or $1$). Typically, $M$ is the sampling pattern dictated by the LiDAR and is usually agnostic of the scene.

We wish to design a sampling network $g$, termed SampleDepth, that uses some prior $\hat{D}^{-}$ on the scene to produce a better sampling mask $M$ with the same number of samples $k$:  
\begin{equation} \label{eq:*} 
M^* = \underset{M}{argmin}||f(D \odot g(\hat{D}^{-}),I) - D||_2^{2}~~~\mbox{s.t.} |M| = k 
\end{equation}

As proposed by Dai~\etal~\cite{dai2019adaptive}, we use a simple yet efficient method for sampling 2D inputs. In essence, the principle involves creating a mask and then element-wise multiplying the input signal with it. 

Figure~\ref{fig:training-regime} gives an overview of our proposed approach. First, the prior ${\hat D}^{-}$ is given as input to $\mathbb{U}(.)$, a convolution neural network (CNN) based on U-Net~\cite{ronneberger2015u} architecture. The output of this network is a probability map, $P_{ro}\in\mathbb{R}^{H{\times}W{\times}2}$. The channels of $P_{ro}$ denote the probability of the pixels to be sampled. Next, by applying SoftArgmax~\cite{finn2016deep,goroshin2015learning}, a differentiable operator, on the probability map $P_{ro}$, the sampling mask $M\in\mathbb{R}^{H{\times}W}$ is created. This mask is a close approximation of a binary mask. Finally, the sampling mask is multiplied, element-wise, with the ground truth depth map $D$, in order to obtain the sampled depth map $D_S$.

The basic training regime of SampleDepth consists of three steps. First, a depth completion task is pre-trained independently with an agnostically  sampled signal input and an RGB image. Then, we freeze the weights of the task network. Next, SampleDepth takes the signal and samples it. The output of SampleDepth is then fed into the frozen task network. Finally, the task network and SampleDepth are fine-tuned together. 

In order to train our model, we use the following loss terms:
\begin{equation} \label{eq:*} 
 L_{total} = L_{task}(D_s,I) + {\alpha}L_{sample}(M)
\end{equation} 
The first term, $L_{task}(D_s,I)$, optimizes the sampled depth map $D_s$ to the task. It preserves the task performance with the sample depth map. $L_{sample}(M)$ is responsible to reduce the number of sampled points to the desired value. This can be formulated as: 
\begin{equation} \label{eq:*} 
 L_{sample}(M) = \dfrac{||M| -k|}{k}
\end{equation} 
Where $k$ represents the desired number of sampled points. 

\begin{figure}[b]
    \centering
    \includegraphics[width=\linewidth]{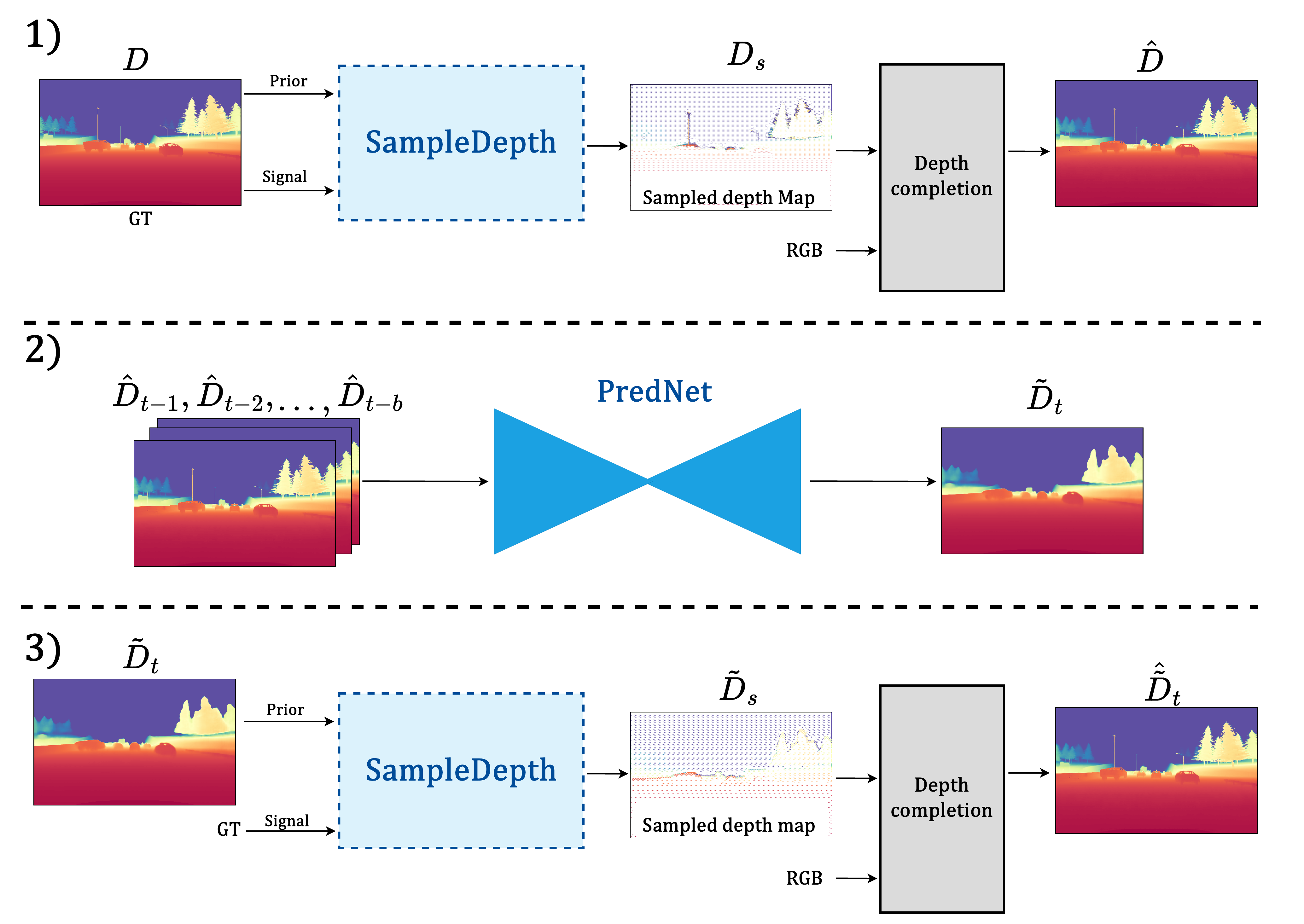}
    \caption{\textbf{Training regime of the proposed Adaptive Model}. This figure refers to the \textbf{PredNet} version. In the \textbf{ImplicitPred} model stage 2 is not performed. Instead, The previously constructed depth maps, $\hat{D}_{t-1:t-b}$, are directly fed into SampleDepth in stage 3.}
    \label{fig:PredNet_train}
\end{figure}

\subsection{An Adaptive Model } \label{TowardsPrior}
For applications such as adaptive LiDAR, we aim to sample the depth map $D_t$, at time $t$, given past reconstructed depth maps ${\hat D}^{-}_{t-1:t-b} \triangleq \{{\hat D}_{t-1},{\hat D}_{t-1},...,{\hat D}_{t-b}\}$ , where $b{\geq}1$. We evaluated two variants of an adaptive model:

\paragraph{PredNet.} In this variant, an explicit prediction of the current reconstructed depth map, $\tilde{D_t}$, is created based on past reconstructed maps, $\hat{D}_{t-1:t-b}$. The predicted reconstructed depth map, $\tilde{D_t}$, serves as the prior to sample the real-world scene, i.e., the ground-truth (GT).

The full training regime of the PredNet process is depicted in Figure~\ref{fig:PredNet_train}. First, SampleDepth is trained with $D$, GT depth maps, which served both as a signal and as a prior. The reconstructed depth maps, $\hat{D}_{t}$, estimated during training, are saved. This will be used as the past reconstructed maps, $\hat{D}_{t-1:t-b}$, in the next step.

Next, in order to create the predicted reconstructed map, $\Tilde{D}_t$, the past reconstructed maps, $\hat{D}_{t-1:t-b}$ are given as input to a U-Net based network (PredNet). The PredNet loss is an $L_1$ with respect to the GT. 

Finally, the explicit prediction of the current reconstructed depth map, $\Tilde{D}_t$, is being used as a prior to sampling the depth from the real-world scene.

\paragraph{ImplicitPred.} This variant bypass the depth map reconstruction step altogether. Instead, past reconstructed depth maps, $\hat{D}_{t-1:t-b}$, are fed directly to SampleDepth as priors.

In this case, an additional loss term, $L_{SampledMaps}$ has been included. This loss term encourages the depth maps generated at this stage, $\hat{D}_s$, to be as close as possible to the sampled depth map, $D_s$, generated where GT serves as both the signal and the prior. For this purpose, $L_1$ loss is used.

\subsection{An End-to-End Adaptive Model} \label{End2End_Adaptive_Model}
We trained PredNet using $\hat{D}_{t-1:t-b}$, as described in subsection~\ref{TowardsPrior}. But this data was created using GT priors, which is not available in reality. To provide a complete solution, we would like to remove this constraint. The reconstructed depths maps should be derived from previous iterations of the algorithm, in which the GT is not used as a prior. Our End-to-End solution is summarized in Algorithm~\ref{alg:PseudoEndtoEnd}.

\begin{algorithm}
\caption{End-to-End Adaptive Model}\label{alg:PseudoEndtoEnd}
\begin{algorithmic}

\State \textbf{Inputs}: Images $I_{1:T}$, Dense Depth Maps $D_{1:T}$, Memory size $S$, Number sampled points $k$
\State {$b \gets S$}
\For {$t \gets 1$ to $T$}
    \If{$t< S$}
        \State {${D}_s \gets SampleRandom(D_t,k) $}
    \Else 
        \State {$\tilde{D}_t \gets PredNet(\hat{D}_{t-1:t-b})$} 
        \State {${D}_s \gets SampleDepth_k(\tilde{D}_t)\odot D_t$} 
    \EndIf

\State {$\hat{D}_t \gets DepthCompletion({D}_s,I_t)$}

\EndFor
\end{algorithmic}
\end{algorithm}

To clarify, in the step in which SampleDepth is being used, the depth map, $D_{t}$, refers to the signal that is being sampled, and the reconstructed map, $\tilde{D}_t$, refers to the prior. In addition, $SampleDepth_k$ refers to a network that was trained to sample $k$ points.

\section{Experiments}

\bgroup
\def\arraystretch{1.3}%
 
 \begin{table*}[bth!]
 \begin{center}
 \begin{adjustbox}{width=1\textwidth}
 \begin{tabular}{l | c c | c c c}
\toprule


& & & \multicolumn{2}{c}{$19K$ sampled points} &  \multicolumn{1}{c}{$4.8K$ sampled points} \\ 
\cmidrule(lr){4-5}
\cmidrule(lr){6-6}

\textbf{Method} & \textbf{Sub-method} & \textbf{Prior}  & \textbf{SHIFT (dense)} & \textbf{Pseudo KITTI (dense) }&\textbf{KITTI (sparse)}   \\

\hline
\hline
\textbf{Agnostic}& - & - & 3.158 & 0.825 & 0.787 \\ 

\cline{1-3}
\multirow{4}{*}{\textbf{SampleDepth} }& Fixed mask & Training set & 2.974 & 0.681 & 0.672\\

\cline{2-3}
&Adaptive (ImplicitPred) & Previous reconstructed depth maps & 2.525 & - & -\\
\cline{2-3}
&Adaptive (PredNet) & Previous reconstructed depth maps & 2.472 & 0.652 & - \\
\cline{2-3}
& Lower bound & Sampled signal(GT depth) & 1.643 & 0.631 & 0.52 \\

\bottomrule

\end{tabular}
\end{adjustbox}
\end{center}

\caption{\textbf{Comparison of different sampling methods on two datasets (SHIFT, KITTI)}. In both datasets, the adaptive method outperforms the agnostic and fixed mask methods. The SHIFT dataset contains dense depth maps that have been sampled. Pseudo KITTI (dense) refers to the pseudo GT which was created and sampled. KITTI (sparse) denotes that the original GT has been sampled. See text for details. Results are reported in the form of RMSE[m].}
\label{tab:priorComp}
\end{table*}
\egroup

\subsection{Datasets and Implementation Details}
We conduct experiments on two datasets:  KITTI Depth Completion \cite{uhrig2017sparsity}\cite{geiger2012we} and SHIFT\cite{sun2022shift}. 

\textbf{KITTI Depth Completion:} The dataset provides both RGB images and aligned sparse depth maps generated by projecting 3D LiDAR points on the corresponding image frames. A sparse LiDAR map has about $5\%$ valid pixels and a ground truth depth map has around $16\%$ valid pixels. The dataset contains $86K$ frames for training, together with $7K$ validation frames and $1K$ test frames. The validation has a split of $1K$ subset frames, which is referred to as "selected validation set". Following Gansbeke~\etal~\cite{van2019sparse}, for both training and testing, input images and depth maps were cropped to $256\times 1216$ due to the lack of LiDAR points in the top part of the images. 

\textbf{SHIFT:} is a synthetic dataset for autonomous driving. It is designed to provide highly diverse environmental conditions for real-world applications. We used the official discrete split that was released by the authors. A total of $3K$ training sequences and $500$ validation sequences are included. Each sequence is composed of $51$ frames, which are sampled at $1Hz$. We have down-scaled the images to a size of $400 \times 640$ pixels.

\paragraph{Implementation Details.} 
SampleDepth is implemented in PyTorch~\cite{paszke2017automatic} and trained on four NVIDIA GTX 1080 Ti and evaluated using one GPU. As our depth completion basline we used the network proposed by Gansbeke~\etal\cite{van2019sparse}. Li~\etal~\cite{li2020multi} depth completion network was also used for further ablation studies. A detailed description of the training regime can be found in the supplementary. 

\paragraph{Evaluation Method.}
SampleDepth is trained to optimize a depth completion downstream task. Therefore, the proposed method is evaluated with respect to the quality of the reconstructed depth maps. As is common, we use the root mean square error (RMSE) as well as the mean absolute error (MAE). 

SampleDepth is trained to produce a pre-defined number of sampling points. This works well with the SHIFT dataset because it is synthetic. That is, for every sampled pixel there is a corresponding depth value.

This is not the case in the KITTI dataset because the depth data is sparse. This is true even if we use the ground truth depth map provided in the KITTI dataset. In order to address this issue, we propose two approaches:

\textbf{KITTI (sparse).} We sample more pixels than is actually required and then use only pixels that actually have a depth value associated with them.

\textbf{Pseudo KITTI (dense).} As part of the pre-processing step, we trained Gansbeke~\etal\cite{van2019sparse} to estimate a dense depth map by randomly selecting $20\%$ points from the GT. These dense depth maps were saved and used for pseudo GT purposes.

\paragraph{Fixed mask.}
As a baseline above the {\em agnositc} method being used today, we compute a fixed mask. This method is intended to learn a single fixed sampling mask that is optimized for the entire data set. For this purpose, we use SampleDepth without any prior input. The U-Net encoder-decoder was removed and we initialized the probability map, $P_{ro}$, values with kaiming normal~\cite{he2015delving}. The {\em fixed} mask was then used as-is during inference to sample the GT.

\begin{figure}[bt]
    \centering
    \includegraphics[trim=0.2cm 24.5cm 0.25cm 0cm, width=\linewidth]{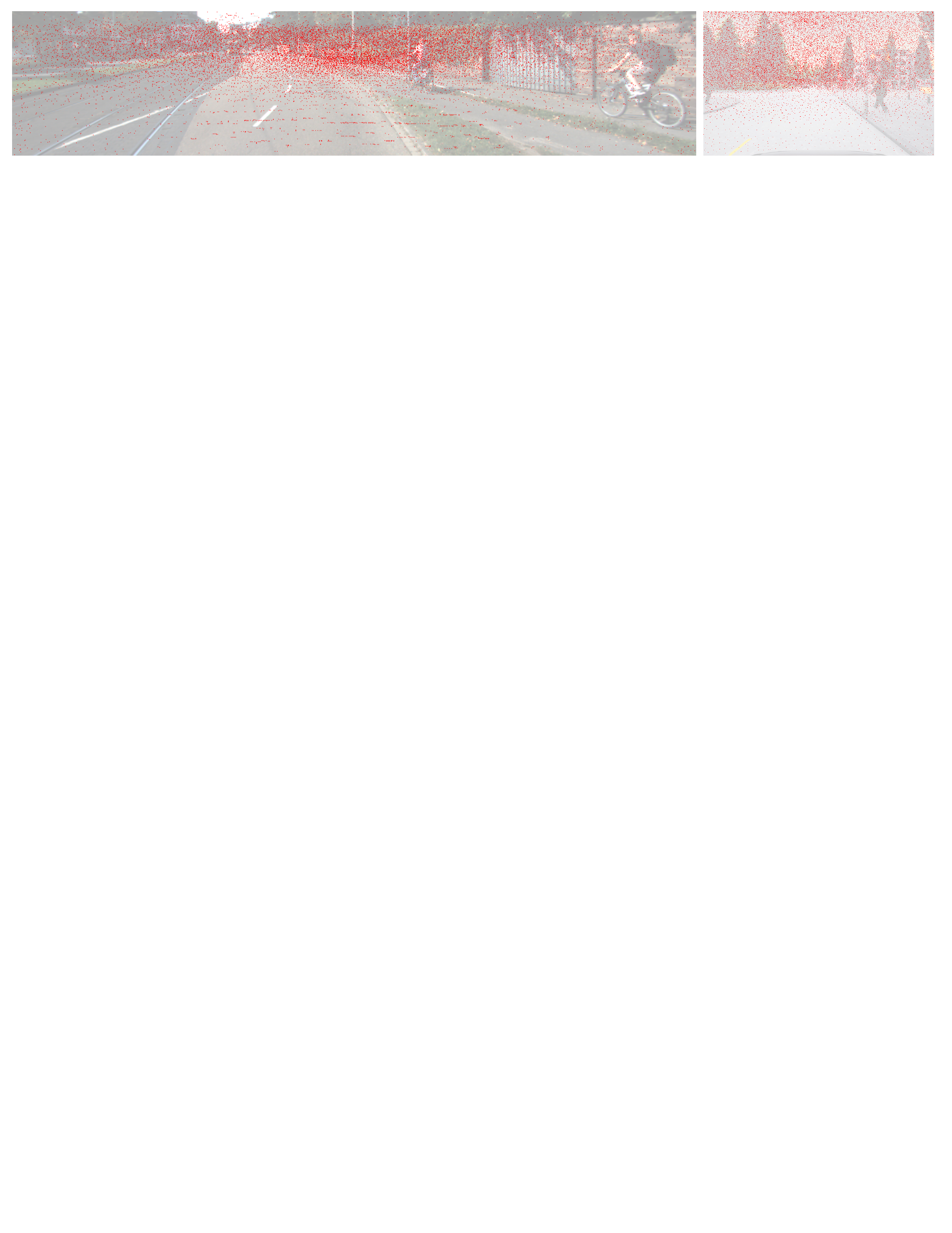}
    \caption{\textbf{Fixed sampling mask.} Two examples of the fixed mask (red dots) on images from the datasets. The mask contains roughly $19K$ points. (Left) KITTI dataset. (Right) SHIFT dataset. In both cases, the majority of the sampling points is concentrated in the upper half of the mask. Objects in these areas are typically located at greater distances.}
    \label{fig:global_mask}
\end{figure}

Figure~\ref{fig:global_mask} shows the fixed sampling map for the KITTI and SHIFT datasets. As can be seen, the sampling pattern are not identical and exhibit the difference between the two datasets. We observe that a greater number of points are sampled from the upper half of the image for both datasets. Due to the automotive structure of the scene, the objects in these parts are typically located at a greater distance.

\subsection{Results}\label{results}
Table~\ref{tab:priorComp} reports results of our method in different settings and on different datasets.

The top part of the table shows results of sampling $19$K points in the SHIFT dataset. As can be seen, the agnostic approach achieves a depth completion RMSE of $3.158$m. A fixed mask improves the error to $2.974$m, while adaptive methods cut the error down to $2.472$m. The "Lower bound" row shows the error if we had access to the ground truth as a prior. In this case, the error is $1.643$m.

ImplictPred uses the previously reconstructed depth maps as a prior to predict where to sample. In PredNet, a predication of the current depth map is made first, and then serves as a prior to the sampling of the current time step depth map.
We found that using the last four frames gave the best results for PredNet and the last two frames gave the best results for ImplictPred.

The two adaptive methods are compared in Table~\ref{tab:priorComp} for $19K$ sampled points. The two methods yield very similar results, with PredNet performing slightly better. Compared with agnostic sampling, PredNet shows a significant improvement of $22\%$, while ImplictPred gains $20\%$ over agnostic sampling. Similar results can be observed in the KITTI dataset when using either the sparse depth maps, or the completed depth maps, as the ground truth to sample from. We conclude that adaptive sampling does indeed reduce the RMSE error of the reconstructed depth maps.

\begin{figure*}[thb!]
    \centering
    \includegraphics[trim=0.8cm 21.6cm 0.8cm 0.3cm,width=17.2cm]{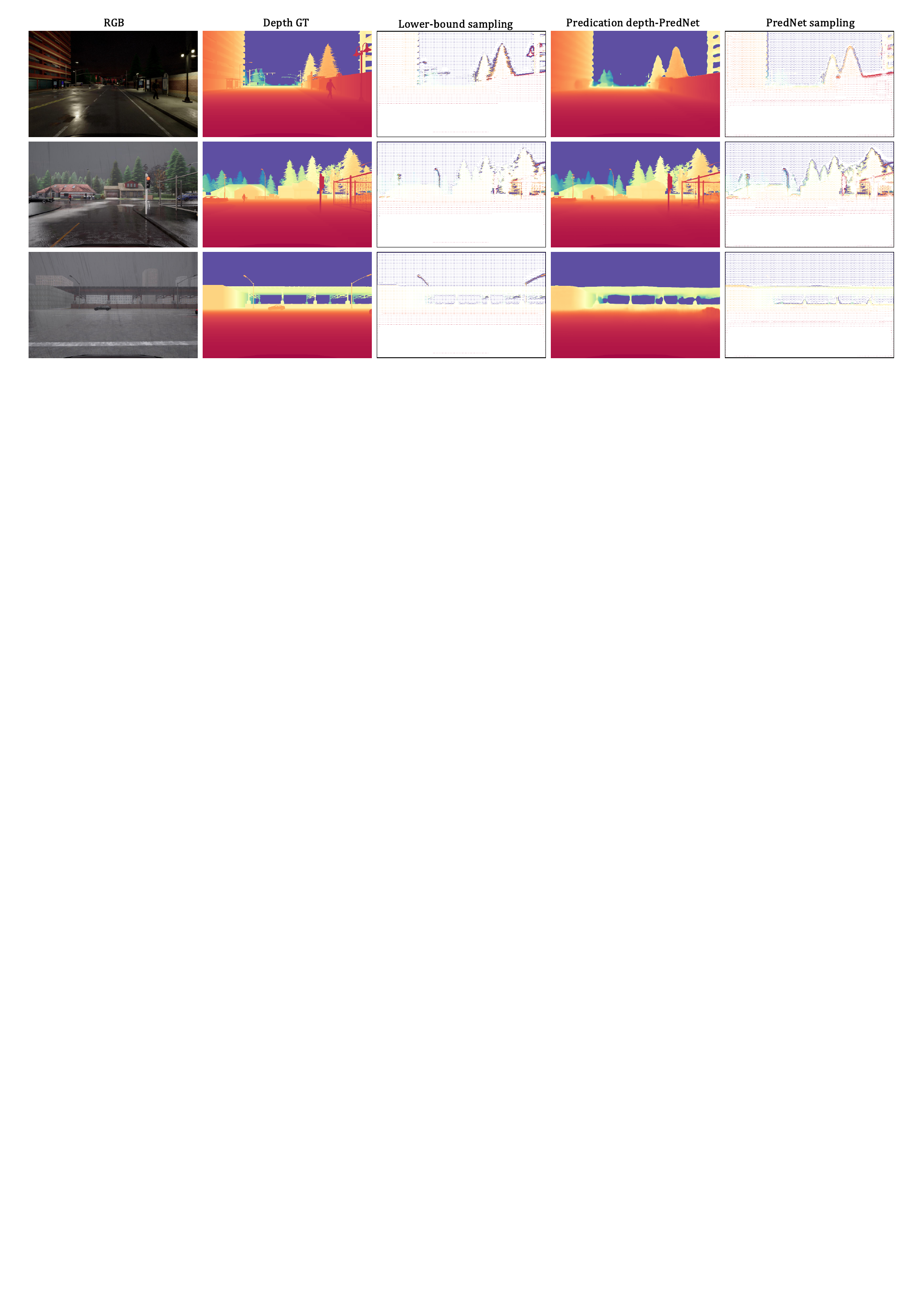}
    \caption{\textbf{Lower Bound vs PredNet sampling.} We compare two priors to sampling the GT. PredNet is using previous frames as prior. Lower Bound is a theoretical approach, where the GT itself is used as a prior. From left to right we show the input RGB image, the Depth GT that is to be sampled, the sampling mask based on the GT itself (this forms a lower-bound on sampling), the predicted depth map by our PredNet, and finally the PredNet based sampling mask.
    The sampling mask is less dense in regions where the prediction maps fails to predict fine details. For example, in the first and third rows, PredNet did not predict the traffic signs and light poles, therefor the correlated sample maps have a lower density in those regions.   
    }

    \label{fig:Prednet_vs_Lowerbound_vis}
\end{figure*}

A quantitative comparison of sampled patterns between the Lower-bound and PredNet variants is presented in Figure~\ref{fig:Prednet_vs_Lowerbound_vis}. There is a tendency for both methods to sample fewer points in very close depth ranges. In the Lower-bound sampling pattern, there is a high concentration of sampling points near the edges. In PredNet, in certain cases, the edge patterns appear more blurry, while in other cases, some of the edge details have almost completely disappeared.

\begin{figure}[t]
    \centering
    \includegraphics[width=\linewidth]{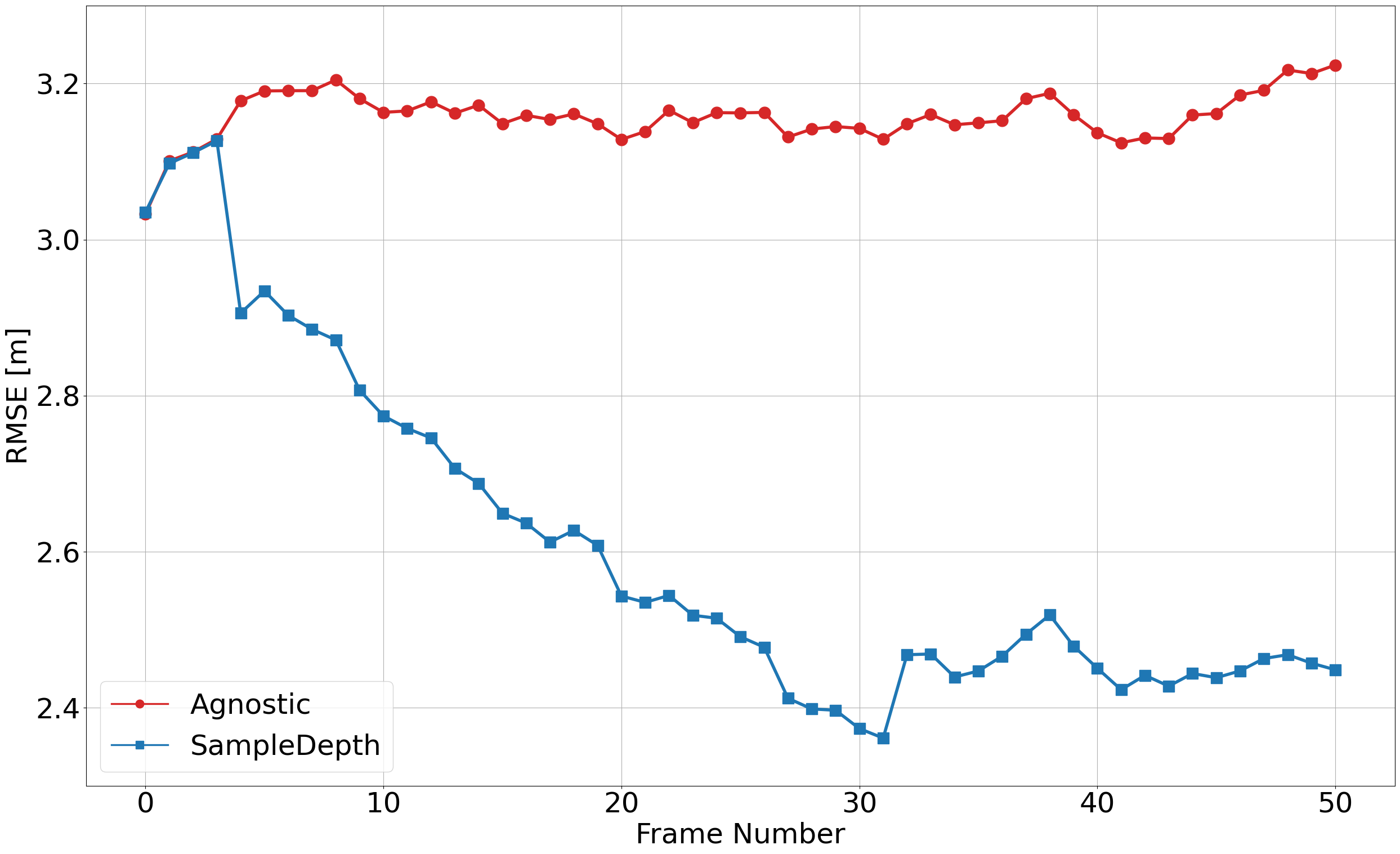}
    \caption{\textbf{SampleDepth End-to-End adaptive model solution mean accuracy per frame.} SHIFT discrete validation set has 500 different sequence of driving scenes. At 1Hz each sequence has 51 frames.}
    \label{fig:end_to_end_per_frame}
\end{figure}

\paragraph{End-to-End Adaptive Model} 
Putting all the pieces together, we evaluate the performance of the adaptive method on a longer sequences of $51$ frames. We used agnostic sampling on the first four frames of each sequence. After that we used PredNet in order to predict the next reconstructed depth map. This prediction was used as a prior to sample the current time step GT by SampleDepth. The PredNet and SampleDepth networks were not trained or fine-tuned again for this experiment. Weights were taken from the stage described in subsection~\ref{TowardsPrior}.  

Figure~\ref{fig:end_to_end_per_frame} illustrates how well our end-to-end algorithm converges to the same range of accuracy values shown in Table~\ref{tab:priorComp}. The solution remains stable over time and outperforms the agnostic sampling by approximately $25\%$ in the steady state.

 \bgroup
\def\arraystretch{1.5}%

 \begin{table}[thb]
 \begin{adjustbox}{width=\columnwidth,center}

\begin{tabular}{@{\extracolsep{4pt}}lcc*{4}{c@{\enspace}c@{\enspace}c@{\enspace}c@{\enspace}c}}
\toprule
 & \multicolumn{2}{c}{4096} & \multicolumn{2}{c}{2048}& \multicolumn{2}{c}{1024}\\
 \cmidrule{2-3}
 \cmidrule{4-5}
 \cmidrule{6-7}

    {Method} & RMSE & MAE & RMSE & MAE & RMSE & MAE & Inference time \\
   \midrule

   \multirow{1}{*}{\shortstack[c]{Super pixel sampler\cite{wolff2020super}}}
    & - & - & - & - & 1.302 & 0.522 & -  \\

   \multirow{2}{*}{\shortstack[c]{Adaptive LiDAR using \\ Enesmble Variance\cite{gofer2021adaptive} }}
    & \multirow{2}{*}{0.750} & \multirow{2}{*}{0.230} & \multirow{2}{*}{0.900} & \multirow{2}{*}{0.298} & \multirow{2}{*}{1.077} & \multirow{2}{*}{0.473} & \multirow{2}{*}{\shortstack[c]{\textbf{1100ms} \\(2x2080GTX)}}  \\
     & & & & & & &  \\

  \multirow{1}{*}{SampleDepth (Fixed-mask)}
    & {0.747} & {0.245} & {0.959} & {0.304} & {1.140} & {0.420} & \multirow{2}{*}{\shortstack[c]{\textbf{17ms} \\ (1x1080Ti)}} \\

  \multirow{1}{*}{SampleDepth (Lower-bound)}
    & \textbf{0.534} & \textbf{0.181} & \textbf{0.567} & \textbf{0.237} & \textbf{0.751} & \textbf{0.334} & \\
\hline
\hline

  \multirow{2}{*}{\shortstack[c]{SampleDepth(PredNet) \\ (Pseudo KITTI)}}
    &\multirow{2}{*}{1.116} & \multirow{2}{*}{0.425} & \multirow{2}{*}{1.692} & \multirow{2}{*}{0.918} & \multirow{2}{*}{-} & \multirow{2}{*}{-}& \multirow{2}{*}{-} & \\

    &&&&&&&\\

 \bottomrule
\end{tabular}
\end{adjustbox}
\caption{\textbf{Quantitative comparison with other methods.} With the exception of SampleDepth(PredNet), all comparisons were made using the KITTI Depth Completion validation set, which contains sparse GT.}
\label{tab:sota_comparison}
\end{table}
\egroup


\paragraph{Comparison with the state-of-the-art}
The closest alternatives we could find in the literature are the work of 
Wolf~\etal\cite{wolff2020super} and Gofer \etal~\cite{gofer2021adaptive} that represent the current state-of-the-art depth sampling on the KITTI Depth Completion dataset. 

A fair comparison between these methods and ours is not possible, because they do not use prior depth maps and we do. On the other hand, our SampleDepth does not use the RGB image as input, while they do. Keeping this in mind, we use the sparse GT provided by KITTI as the signal to sample from, and report results in Table~\ref{tab:sota_comparison}.

We evaluate our method and that of Gofer~\etal~\cite{gofer2021adaptive} on three sampling rates and observe that SampleDepth (The Lower-bound version) outperforms Gofer \etal~\cite{gofer2021adaptive} in all cases. Moreover, even our {\em fixed} solution, which does not take any prior during inference, and generates only one mask across the entire dataset, is able to achieve similar performance to Gofer~\etal~\cite{gofer2021adaptive}. {\em Fixed}-mask performance is slightly better for $4K$ sampled points, but slightly worse for lower sample rates.
Additionally, our method offers a much faster inference time, that can be executed on edge devices. We also report in the table the results of Wolf~\etal~\cite{wolff2020super} for the lowest sampling ratio of only 1,024 points and observe that our method outperforms it.

Two other methods, that sample depth maps for depth completion tasks, are not mentioned in the table. The method of Tcenov \etal~\cite{tcenov2022guide} is suitable for sampling only a dense signal. Therefore, their results are reported only on the simulated dataset Synthia~\etal\cite{ros2016synthia}. Bergman~\cite{bergman2020deep} examined only very low sampling rates (fewer than $1000$ points).

\begin{figure}[b]
    \centering
    \includegraphics[width=\linewidth]{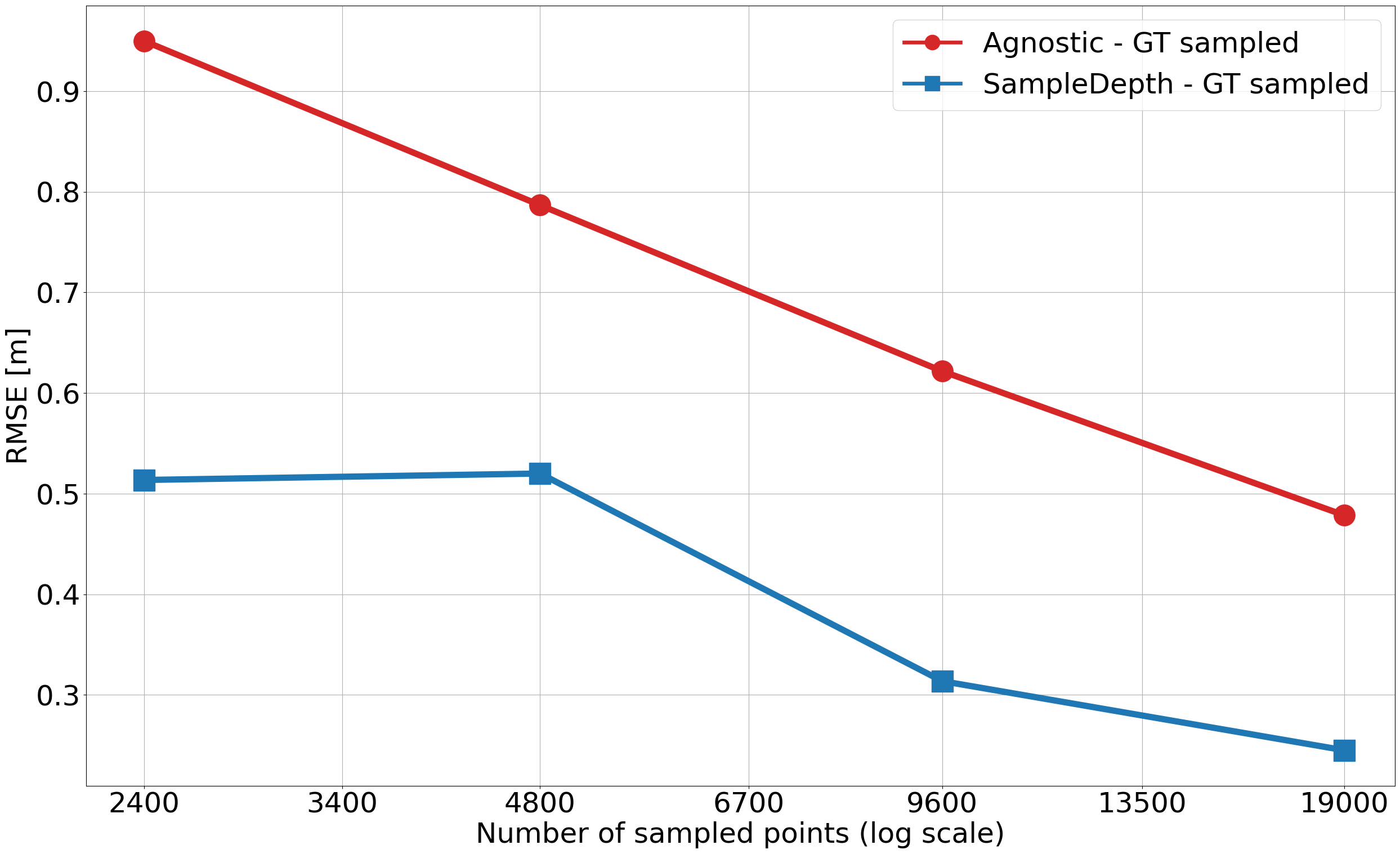}
    \caption{\textbf{Depth completion accuracy with SampleDepth.} Regarding the sampling process in this graph, the sampled signal is the prior itself. KITTI Depth Completion selected validation set is used for evaluation.}
    \label{fig:kitti_LiDAR_and_GT_sample_ratio}
\end{figure}

\subsection{Ablation Study}\label{ablation}

\begin{figure}[bt]
    \centering
    \includegraphics[trim=0cm 14cm 0cm 0cm, width=\linewidth]{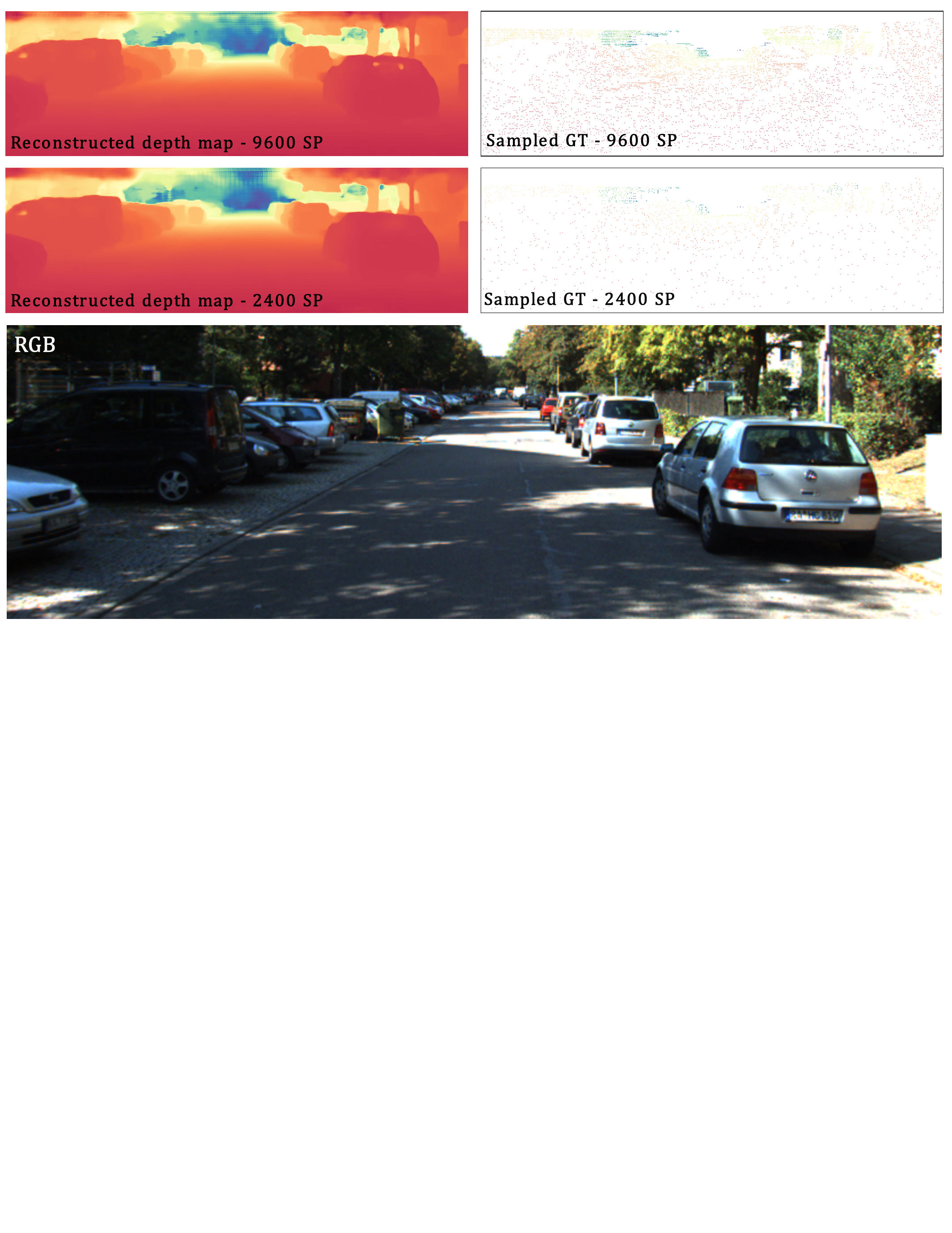}
    \caption{\textbf{KITTI GT sampling using SampldeDepth.} Two sampled points selected, 2400 and 9600, and sampled using the Lower-bound method. Resulting depth maps from both reconstructions are very similar.}
    \label{fig:kitti_gt_sample}
\end{figure}

\paragraph{Sampling Ratio}

Figure~\ref{fig:kitti_LiDAR_and_GT_sample_ratio} presents a comparison between agnostic sampling and SampleDepth for various sampling ratios. In this case, we are sampling from the ground truth depth maps provided in the KITTI dataset. The original GT depth maps consist of approximately $90K$ points. SampleDepth leads to better depth completion results for all tested sampling ratios, from $19K$, all the way down to 2,400 points (which is equivalent to a sampling rate of 1:37)

Figure~\ref{fig:kitti_gt_sample} illustrates the quantitative results obtained from sampling the GT signal using SampleDepth for 2,400 and 9,600 points. As can be seen, the completed depth map in both cases is almost indistinguishable. We provide additional results in the supplementary material in which LiDAR signals are being sampled. 

\begin{figure}[b]
    \centering
    \includegraphics[width=\linewidth]{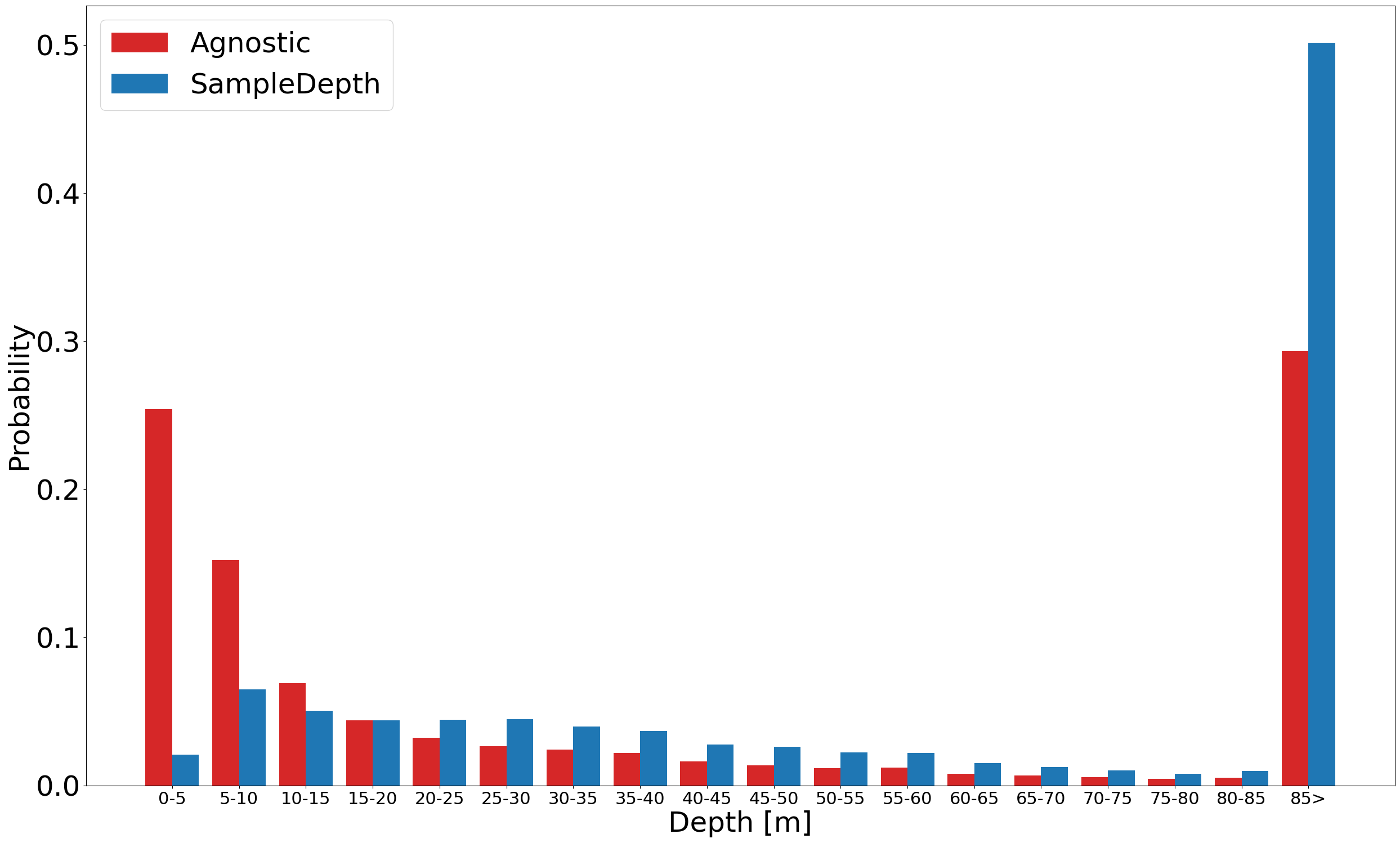}
    \caption{\textbf{Sampled depth distribution.} The sample distribution is derived from evaluation of the agnsotic and ours methods on the SHIFT validation dataset. SampleDetph refers to our End-to-End adaptive approach. }
    \label{fig:hist_depth}
\end{figure}

\paragraph {Sampling distribution}
In the quantitative results (Figure~\ref{fig:Prednet_vs_Lowerbound_vis}), we observed that SampldeDepth samples more at far away points. We analyzed this observation further and found that this is indeed the case. Figure~\ref{fig:hist_depth} shows the distribution of depth values at sampled points as collected for both our method and the agnostic approach. SampleDepth has almost twice as many samples at points that are more farther than $85m$. Additionally, there is a significant decrease in the number of points that have been sampled within a distance of less than $10m$. For the intermediate depth ranges, both methods show indistinguishable distribution trends.  A higher percentage of depth samples are taken from $10-15m$, and the number of sampled points decreases as the depth is increased.

\paragraph{Transferability} SampleDepth can be trained with different depth completion networks. In this experiment, we use the depth completion networks proposed by Gansbeke \etal~\cite{van2019sparse} and Li \etal~\cite{li2020multi}. Table~\ref{tab:several_downstram_depth_comp} compares SampleDepth, trained with two different downstream tasks, against the agnostic method. As can be seen, using SampleDepth cuts RMSE error by about $50\%$, and the MAE by about $25\%$, compared to the agnostic method, for both downstream tasks.

In fact, it is possible to train SampleDepth using one depth completion network and use it, as is, with another network, as can be seen in Table~\ref{tbl:mixed_and_match}. In the experiments reported in this table, the depth completion network is kept frozen and only SampleDepth is trained. Observe that in the third row of the table SampleDepth was trained on the network of Gansbeke \etal~\cite{van2019sparse} and then combined with the network of Li \etal~\cite{li2020multi} with no further finetuning. The results are nearly indistinguishable from the regular case where we train and test SampleDepth with the same depth completion network (the top two rows in the table).

 \begin{table}[tb]
 \begin{adjustbox}{width=\columnwidth,center}

\begin{tabular}{@{\extracolsep{4pt}}lcc*{4}{c@{\enspace}c@{\enspace}c@{\enspace}c@{\enspace}c}}
\toprule
 & \multicolumn{2}{c}{Agnostic} & \multicolumn{2}{c}{SampleDepth}\\
 \cmidrule{2-3}
 \cmidrule{4-5}
 \cmidrule{6-7}

    {Downstream network} & RMSE & MAE & RMSE & MAE  \\
   \midrule
   \hline

   \multirow{1}{*}{Gansbeke~\etal\cite{van2019sparse}}
    & 0.4785 & 0.118 & \textbf{0.245} & \textbf{0.097}   \\
\addlinespace
\addlinespace

\addlinespace
  \multirow{1}{*}{Li~\etal\cite{li2020multi}}
    & 0.495 & 0.126 & \textbf{0.254} & \textbf{0.089} \\
 \bottomrule
\end{tabular}
\end{adjustbox}
\caption{\textbf{A comparison of our solution vs agnostic sampling with two downstream depth completion networks.} Here we use the "Lower bound" sampling method for SampleDepth. The GT depth maps have been sampled to $19K$ points and have been evaluated on KITTI Depth Completion selected validation set.}

\label{tab:several_downstram_depth_comp}
\end{table}

\bgroup
\def\arraystretch{1.5}
 \begin{table}[tb]
 \begin{adjustbox}{width=\columnwidth,center}
\begin{tabular}{l| c | c |c  c| c  }
\toprule
Mode & SampleDepth  &Downstream task & RMSE & MAE & SP  \\
\hline
\hline
Regular& {Li~\etal\cite{li2020multi}} & {Li~\etal\cite{li2020multi}} & 0.327 &  0.113 & $17.1$k\\
Regular &Gansbeke~\etal\cite{van2019sparse} & Gansbeke~\etal\cite{van2019sparse} & 0.301 &  \textbf{0.110} & $13.9$k\\
\hline
Mixed\&Match & Gansbeke~\etal\cite{van2019sparse} &{Li~\etal\cite{li2020multi}} & \textbf{0.293} &  0.119 & $13.9$k\\

\bottomrule
\end{tabular}
\end{adjustbox}
\caption{\textbf{Mixed and match.} In the regular mode, the Downstream task is trained using an agnostic sample of $19K$ points and frozen. Then, SampleDepth is trained while its output is fed into the freeze task network. When using the mixed and match mode, there is no additional training, only  evaluation on the pre-trained mixed Samplers and networks. KITTI Depth Completion selected validation set is used for evaluation. SP stand for sampled points.}
\label{tbl:mixed_and_match}
\end{table}
\egroup

\begin{figure}[t]
    \centering
    \includegraphics[width=\linewidth]{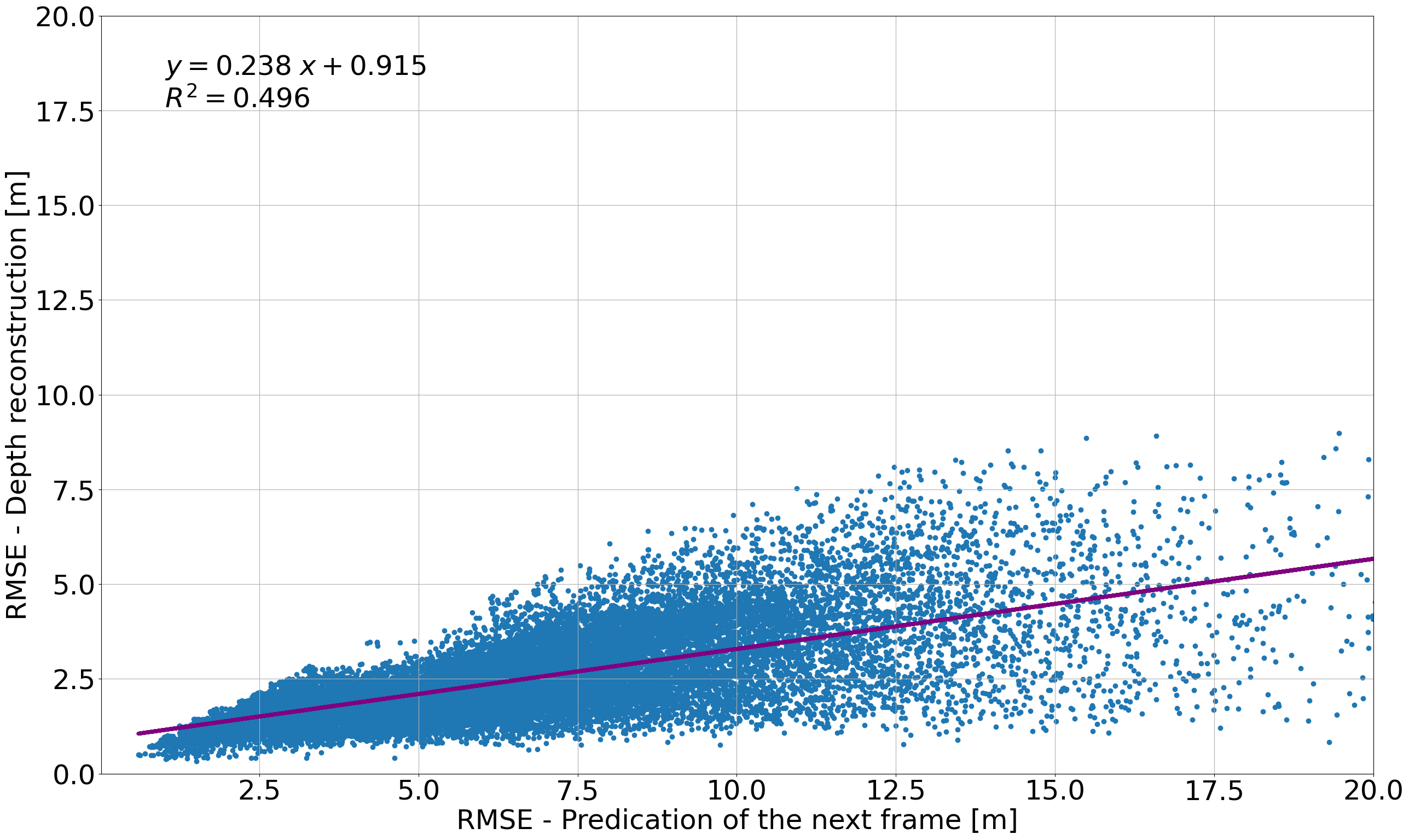}
    \caption{\textbf{Prediction-Completion Correlation.} A better prediction of the next frame will result in a higher quality final reconstruction.}
\label{fig:rec_vs_pred}
\end{figure}

\paragraph{Prediction-Completion Correlation}
Figure~\ref{fig:rec_vs_pred} demonstrate the correlation between the accuracy of predicating the next frame using PredNet, $\Tilde{D}_t$, and the quality of the final reconstructed map $\hat{D}_t$. As can be seen, the better the accuracy of prediction is, the higher the quality of the reconstructed depth map. 

\begin{figure}[tb]
    \centering
    \includegraphics[trim=0cm 11.5cm 0cm 0cm, width=\linewidth]{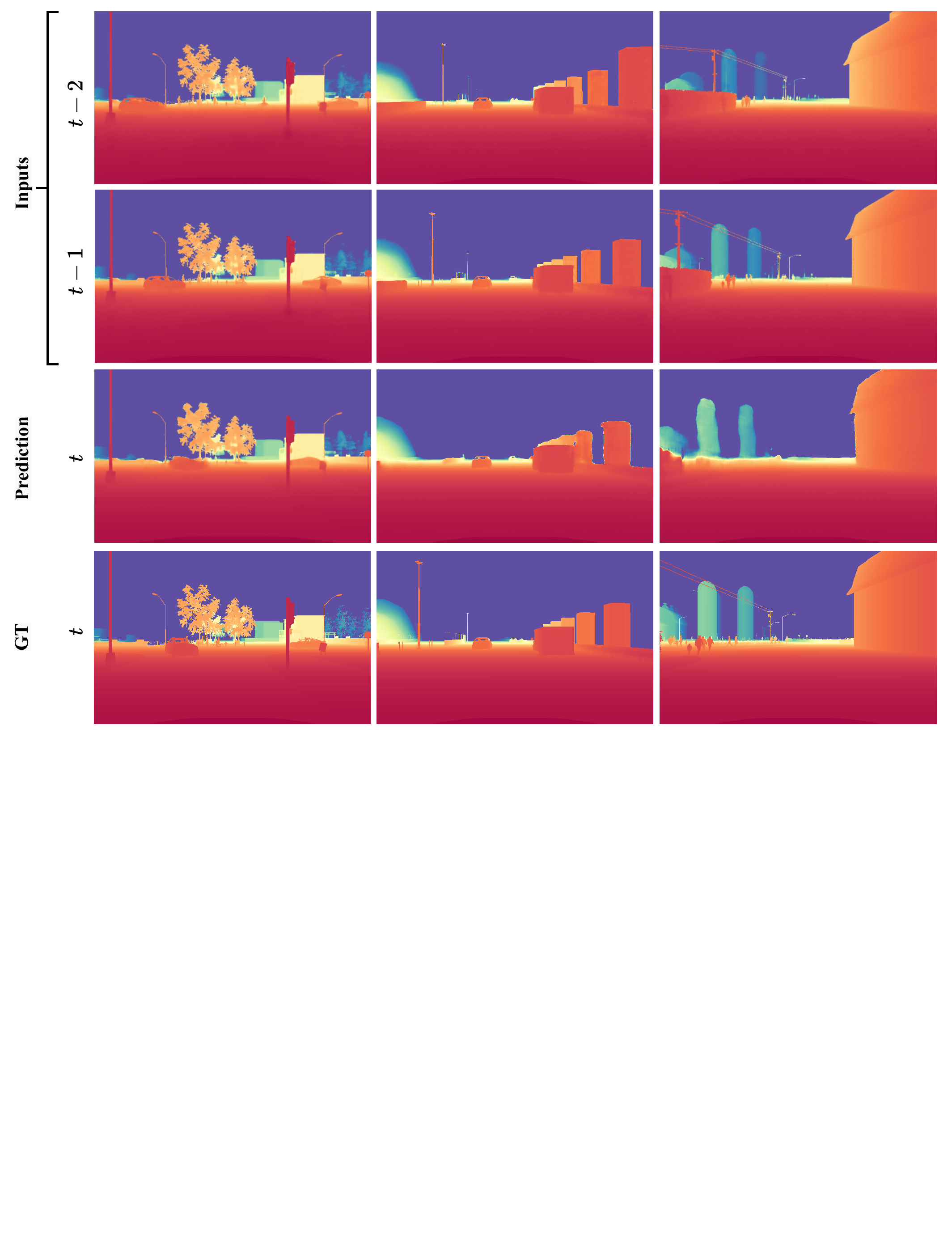}
    \caption{\textbf{PredNet prediction of the next frame.} The network gets K past time reconstructed maps and predict the next frame. Left row: the vehicle in the frame is static, while the other objects in the frame are dynamic. Middle row: driving scenario on a highway. Right row: countryside driving}
    \label{fig:PredNet_prediction}
\end{figure}

\paragraph{Prediction Accuracy}
As can be seen in Figure~\ref{fig:PredNet_prediction}, the larger the ego-motion between successive input frames, $\hat{D}_{t-1:t-b}$, the more difficult it becomes for PredNet to predict the next reconstructed depth map, $\tilde{D}_{t}$. The left column contains sequences in which a vehicle containing the camera is stationary and stops at a red light. PredNet is able to predict the dynamic objects in the scene fairly well, such as the two cars. There are two sequences in the middle and right columns with larger ego-motion between frames. Consequently, PredNet is less effective at predicting fine details.

\section{Conclusions}
We proposed a novel adaptive LiDAR sampling approach, that uses information from previous time steps as a prior in order to predict the current sampling mask. Our solution is based on SampleDepth, a network that optimizes the sampling mask subject to a downstream depth completion task. The use of temporal data allows us to outperform the agnostic sampling mask that is currently being used.
Experiments on two different datasets suggest that our method can reduce depth estimation error by more than $20\%$, given the adaptive sampling masks computed on the fly by SampleDepth. Moreover, we have shown that SampleDepth can be trained with one depth completion network and transferred, as is and with no fine-tuning, to work with a different depth completion network with indistinguishable impact on performance.

{\small
\bibliographystyle{ieee_fullname}
\bibliography{references}
}

\clearpage
\appendix
\section*{Supplementary}

\section{Additional results} \label{sec:supp_additional_results}
\textbf{LiDAR signal sampling.} Certain scenarios, such as memory management or communication, may benefit from the sampling of signals that have already been sampled. This type of application can also be addressed by our SampleDepth solution. 

For this purpose, we sample the LiDAR signal from the KITTI depth completion dataset, while the LiDAR also serves as a prior. 
 
\begin{figure}[thb!]
    \centering
    \includegraphics[width=\linewidth]{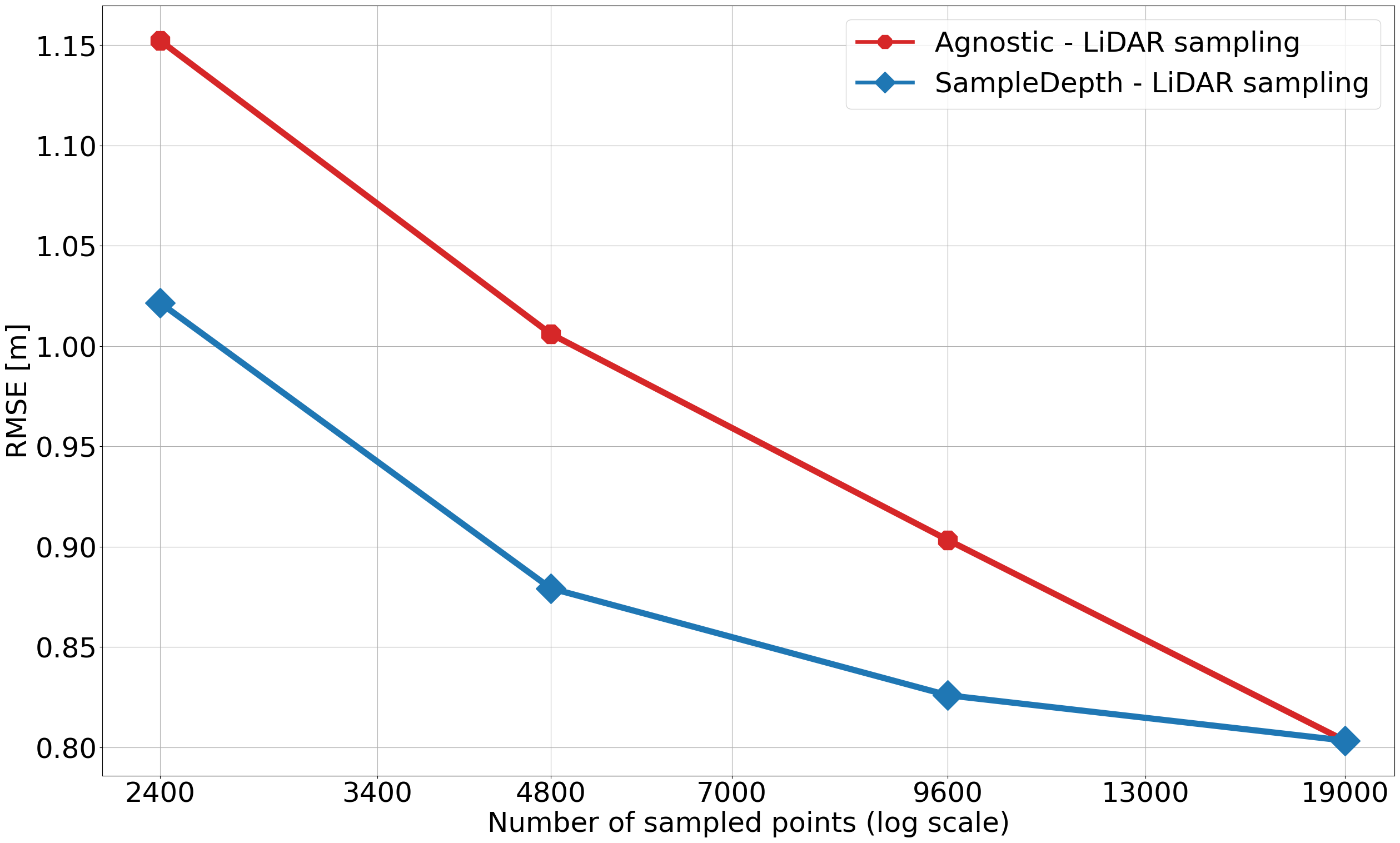}
    \caption{\textbf{LiDAR sampling} Regarding the sampling process in this graph, the sampled signal is the prior itself. KITTI Depth Completion selected validation set is used for evaluation.}
    \label{fig:kitti_LiDAR_sample_ratio}
\end{figure}

\begin{figure}[thb!]
    \centering
    \includegraphics[trim=0.4cm 11cm 0.7cm 1cm, width=\linewidth]{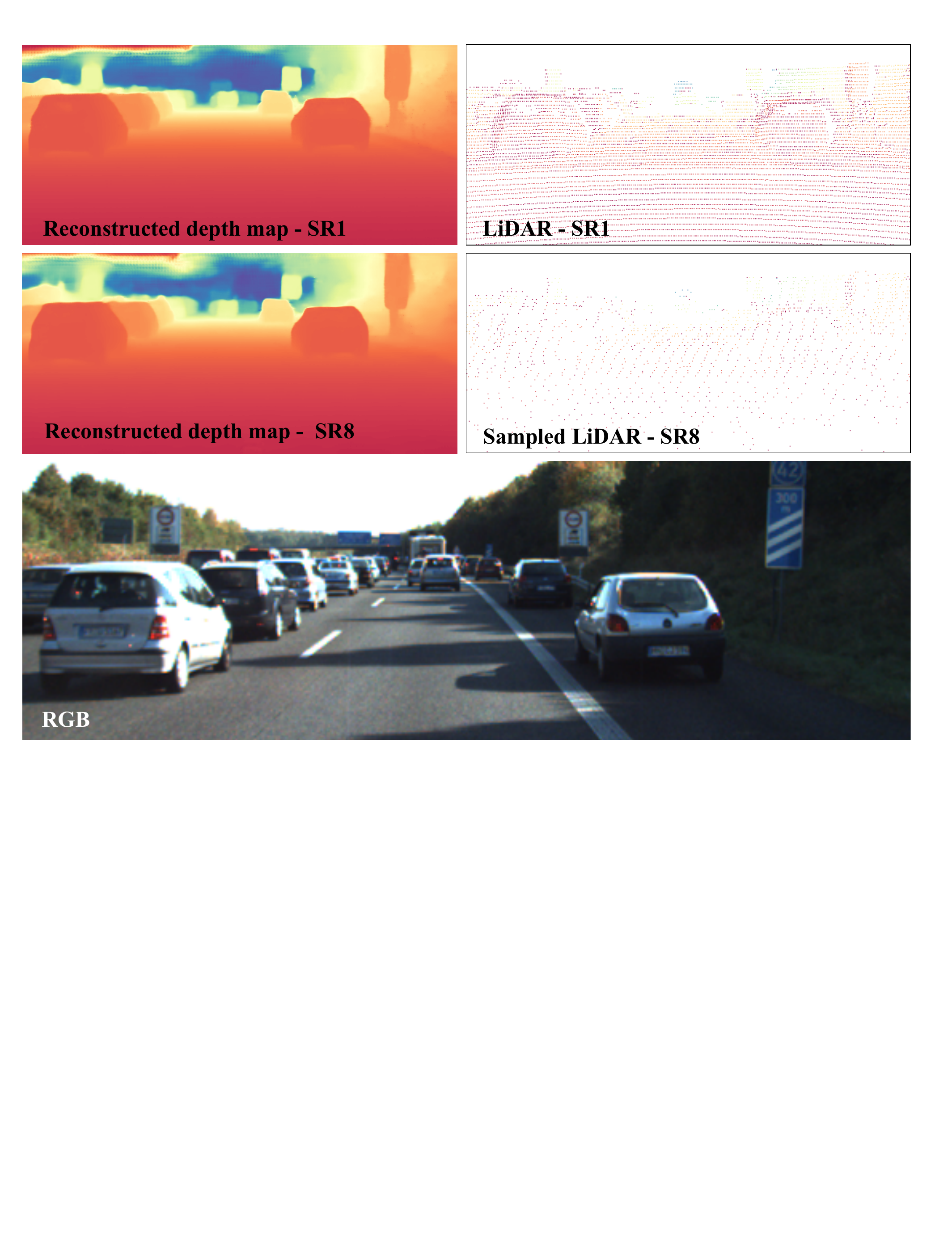}
    \caption{\textbf{KITTI LiDAR Sampling using SampldeDepth.} SampleDepth selects $12.5\%$ of the points from the raw LiDAR. The reconstructed depth maps of SR1 and SR8 are very similar.}
    \label{fig:kitti_lidar_sample}
\end{figure}

 As shown in Figure~\ref{fig:kitti_LiDAR_sample_ratio}, SampleDepth outperforms agnostic sampling for any number of sample points. Figure~\ref{fig:kitti_lidar_sample} illustrates the quantitative results obtained from sampling $12.5\%$ from the LiDAR signal. As can be seen, the depth map generated by the raw LiDAR data is almost identical to the one produced by SampleDepth.

\section{Ablation study}  \label{sec:supp_ablation_study}
\subsection{ImplicitPred}  
In Table~\ref{tbl:ImplicitPred_number_of_frames}, our adaptive solution ImplicitPred is evaluated with different number of past frames, neural network capacity, and regularization. The best performace obatiend with dipper U-Net architecture, include $L_{sampledMaps}$ loss and using 2 past frame. 
\begin{table*}[thb!]
\begin{center}
\begin{tabular}{ l | c |c | c| c c c }
\hline
Architecture  & $L_{task}$& $L_{sample}$ &$L_{SampledMaps}$ & 1 past frames & 2 past frames & 3 past frames  \\
\hline
\hline
U-Net(4 levels) & \checkmark & \checkmark &  & 2.536 & 2.645 & 2.896  \\

U-Net(5 levels) & \checkmark & \checkmark & \checkmark & 2.551 & \textbf{2.525} & 2.542  \\

\hline
\end{tabular}
\end{center}

\caption{{\textbf{ImplicitPred performance as function of architecture, regularization and number of last frames.} The results reported in RMSE[m] for the finial reconstructed depth map. This evaluation was conducted on the SHIFT dataset. The "level" refer to the number of contraction layers. }}
\label{tbl:ImplicitPred_number_of_frames}
\end{table*}

\subsection{PredNet}  
In subsection 4.3 we already demonstrated the correlation between the accuracy of predicating the next frame using PredNet, $\Tilde{D}_t$, and the quality of the final reconstructed map $\hat{D}_t$. 

Based on Table~\ref{tbl:PredNet_number_of_frames}, it can be seen that using four previous frames optimizes the quality of PredNet prediction of the next frame. A greater number of past frames are not tested due to computational limitations. 

\begin{table*}[thb!]
\begin{center}
\begin{tabular}{ l | c c c }
\hline
Architecture  &   2 past frames & 3 past frames & 4 past frames  \\
\hline
\hline

U-Net(5 levels) & 6.845 & 6.742 & \textbf{6.552}  \\

\hline
\end{tabular}
\end{center}

\caption{{\textbf{PredNet performance as function of inputs number of last frames.} The results reported in RMSE[m] for the explicit predication of the next frame. The SHIFT dataset was used for this evaluation. }}
\label{tbl:PredNet_number_of_frames}
\end{table*}

\section{Experimental settings}  
The details of the experiments are presented in Table~\ref{tbl:expirements_details}. Adam~\cite{kingma2014adam} optimizer is used for all experiments. Considering that the table hyper-parameters refer to $19K$ sampled points, it should be possible to work with lower sample ratios.  

The basic training regime of SampleDepth consists of three steps:
First, a depth completion task is pre-trained independently (ID's: a, d, g, j) according to Gansbeke~\etal~\cite{van2019sparse} training regime. Differences between experiments for this stage primarily arise from the depth inputs to the depth completion network. As an example, in the Lower-bound method for 19K sampled points, ID (a), the task network will train while sampling GT in an agnostic manner to $19K$ points.

Then, the task network weights are freeze, and SampleDepth is trained while it's output feed into the frozen task network (ID's: b, e, h, k, t). 

Finally, the task network and SampleDepth are fine-tuned together( ID's: c, f, i, l).

There are some experiments in which weights from other stages are utilized, in order to bypass the three stages. The depth completion task and SampleDepth are loaded with pre-trained weights and have just been fine-tuned.

\begin{table*}[thb!]
\begin{center}
\begin{adjustbox}{width=1\textwidth}
\begin{tabular}{ c | c |c|c c c c c c c }
\hline
Experiment & Dataset & ID & Task network & SampldeDepth & Epochs & Batch size & LR & $\alpha$ & Load weights\\
\hline
    \hline
\multirow{9}{*}{Lower bound}  &           &(a)& Trained & -  & 30 & 18  & 0.001 &-  & *erfnet\cite{van2019sparse}  \\  
            &   KITTI   & (b)&Freeze  & Trained & 20 &  8  &  0.0001 & 4 & (a)\\  
             &           &(c)& Trained & Trained & 7 &  10 &  0.001  & 50 & (b) \\ \cline{2-10}
            &           &(d)& Trained & -  & 30 & 20  & 0.002 &-  &  *erfnet\cite{van2019sparse}  \\  
            &   KITTI Pseudo GT    & (e)&Freeze  & Trained & 20 &  8  &  0.0001 & 25 & (d)\\  
             &           &(f)& Trained & Trained & 10 &  8 &  0.0001  & 50 & (e) \\ \cline{2-10}

            &           & (g)&Trained &    -    & 30 & 20 & 0.008 & - &  *erfnet\cite{van2019sparse}       \\  
            &  SHIFT    &(h)& Freeze  &  Trained & 20 & 10 & 0.0001& 50 & (g)    \\  
            &           &(i)& Trained &  Trained &  7 & 10 & 0.001 & 100 &(h)  \\\cline{1-10}
            
            &           &(j)& Trained & -       & 30 & 18  & 0.001 & - & *erfnet\cite{van2019sparse} \\  
LiDAR       &   KITTI   &(k)& Freeze  & Trained & 20 &  8  &  0.0001 & 4 &(j)\\  
            &           &(l)& Trained & Trained & 7  &  10 &  0.001  & 50 & (k)\\ \cline{1-10}
            
\multirow{3}{*}{Fixed Mask}  &   KITTI    &(m)& Trained & Trained  & 30 & 1  & 0.1 & 50 & (c) \\ \cline{2-10}
            &    KITTI Pseudo GT    &(n)& Trained & Trained  & 30 & 1  & 0.0001 & 5 & (f) \\ \cline{2-10}
            &   SHIFT   &(o)& Trained & Trained   & 30 & 1  & 0.001 & 20& (i) \\ \cline{1-10}

PredNet (Only    &   Pseudo GT KITTI   &(p)& -  & -   & 30 & 10  & 0.0002 & -& -  \\ 
predication of next frame)  &  SHIFT  &(q)& -  & -  & 30 & 10  & 0.001 & - & - \\\cline{1-10}
SampleDepth&   Pseudo GT KITTI &(r) &  Trained  & Trained   & 30 & 10  & 0.0001 & 25& (f) \\ 
(PredNet prior) &  SHIFT    &(s)& Trained  & Trained  & 30 & 10  & 0.001 & 100  & (i)   \\ \cline{1-10}
ImplicitPred &SHIFT   &(t)& Freeze  & Trained   & 30 & 8  & 0.0001 & 100& (i)  \\
\hline
\end{tabular}
\end{adjustbox}
\end{center}
\caption{\textbf{Hyperparameters.} The table provides information on the hyperparameters values used for training SampleDepth at each stage and method.   }
\label{tbl:expirements_details}
\end{table*}

\vfill

\vfill

\end{document}